\documentclass[10pt,twocolumn,letterpaper]{article}

\usepackage{cvpr}
\usepackage{times}
\usepackage{epsfig}
\usepackage{graphicx}
\usepackage{amsmath}
\usepackage{amssymb}
\usepackage{enumitem}
\usepackage{booktabs}
\usepackage{xcolor}
\usepackage{multirow}
\usepackage{comment}
\usepackage{subcaption}
\usepackage{longtable}

\graphicspath{ {figures/}}
\usepackage[pagebackref=true,breaklinks=true,letterpaper=true,colorlinks,bookmarks=false]{hyperref}

\cvprfinalcopy 

\def\cvprPaperID{16} 

\ifcvprfinal\fi
\begin{document}

	\title{Two Stream Self-Supervised Learning for Action Recognition}

\author{Ahmed Taha$^1$ \quad Moustafa Meshry$^1$ \quad Xitong Yang$^1$ \quad Yi-Ting Chen$^2$ \quad Larry Davis$^1$ \\
$^1$University of Maryland, College Park \qquad $^2$Honda Research Institute, USA\\
\tt\small \{ahmdtaha,mmeshry,xyang35\}@cs.umd.edu \quad  lsd@umiacs.umd.edu  \quad ychen@honda-ri.com
}

	\maketitle

\begin{abstract}
	We present a self supervised approach using spatio-temporal signals between video frames for action recognition. A two-stream architecture is leveraged to tangle spatial and temporal representation learning. Our task is formulated as both a sequence verification and spatio-temporal alignment tasks. The former task requires motion temporal structure understanding while the latter couples the learned motion with the spatial representation.
	The self-supervised pre-trained weights effectiveness is validated on the action recognition task.  Quantitative evaluation shows the self-supervised approach competence on three datasets: HMDB51, UCF101, and Honda driving dataset (HDD). Further investigations to boost performance and generalize validity are still required.

\end{abstract}

\section{Introduction}

Big data is a key pillar for the recent deep neural network advancement. Yet even with data availability, annotation costs hinder supervised learning for applications like autonomous car driving. Annotation can be costly for various reasons; it can be time consuming as in action detection. It can also be expensive for requiring expertise as in medical field. In contrast, massive amounts of unannotated videos are untapped. Thus, despite supervised learning success, some domains lag for lacking labeled data. In this paper, we present a self-supervised approach for action recognition.


Recent self-supervised learning frameworks pretrain for an auxiliary task using unlabeled data. Misra et al.~\cite{misra2016shuffle} reason about the frames order in a binary classification setting. To boost performance, Fernando et al.~\cite{fernando2017self} extend the binary classification into a multi-classification problem -- a more difficult setting. In such setting, the network reasons through multiple stacks of frame differences and classifies the odd stack out. To increase task complexity, Lee et al.~\cite{lee2017unsupervised} predict the correct temporal order of frames.

Using the powerful ImageNet weights, spatial image representations are successfully utilized in many applications. In novel domains with scarce annotations, transfer learning is a standard solution. Yet, temporal motion representation still lags due to labeled data scarcity. In this paper, we build on previous work and correlate the self-supervised motion learning task with the successful spatial representation.

\section{Methodology \& Experiments}
\begin{figure}
	\centering
	\includegraphics[width=0.7\linewidth,height=0.9\linewidth]{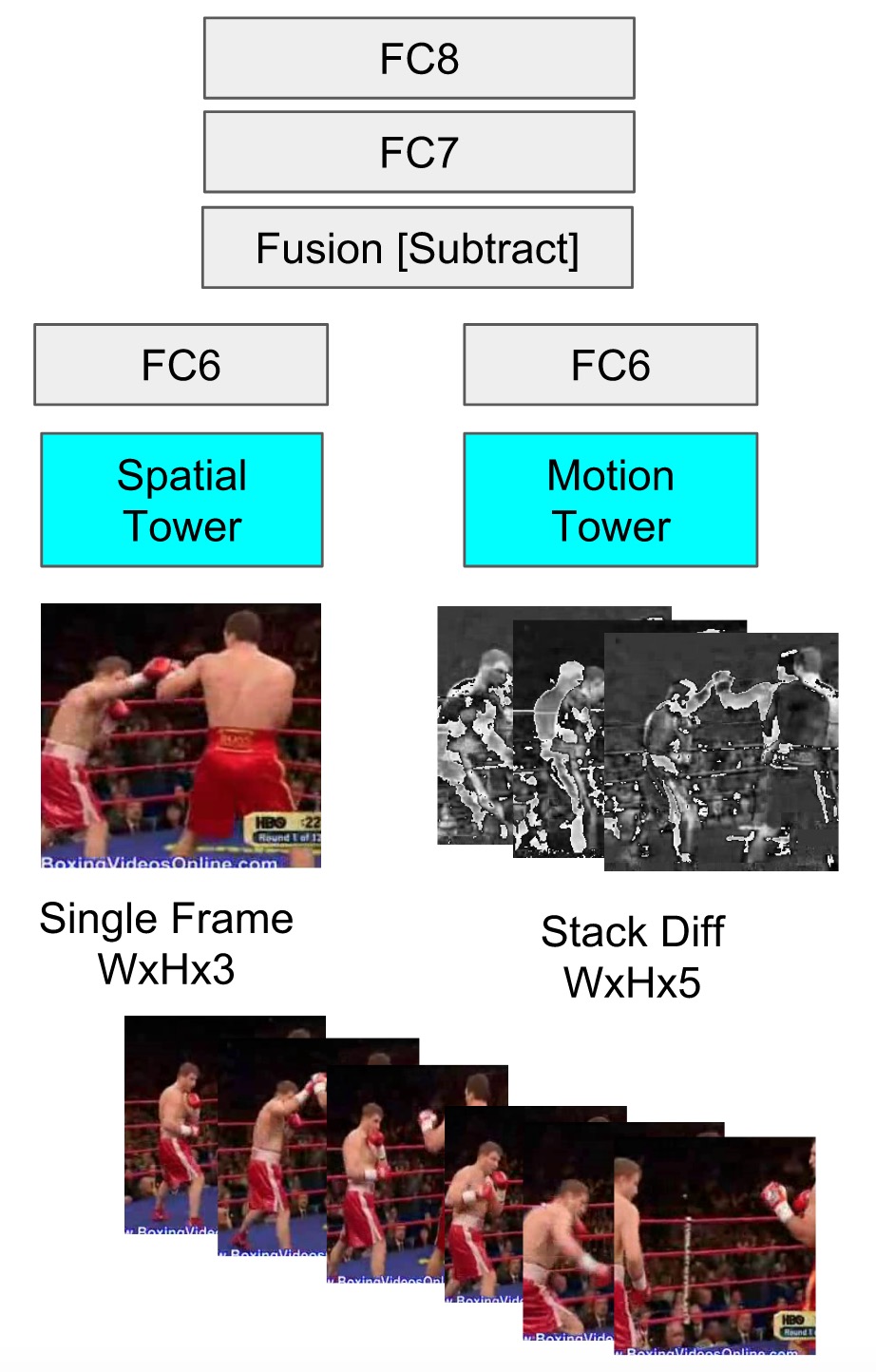}
	\caption{The proposed two stream architecture. Given a clip, the center RGB frame is fed into the spatial tower. Motion, encoded using stack of differences, is fed into the motion tower. Features are fused by subtraction into a spatio-temporal feature after the first fully connected layer --fc6.}
	\label{fig:networkarch}
\end{figure}
\captionsetup[subfigure]{labelformat=empty}
\begin{figure}[t]
	\centering
\begin{subfigure}[b]{0.2\textwidth}
	\includegraphics[width=\textwidth]{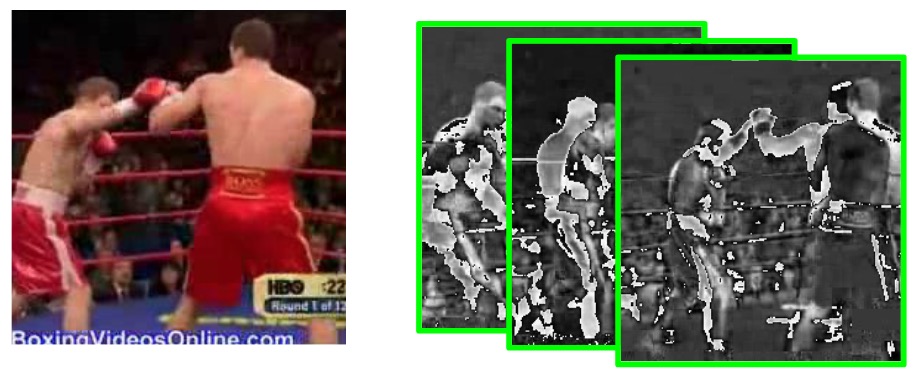}
	\caption{Class I}
	\label{class_1}
\end{subfigure} ~
\begin{subfigure}[b]{0.2\textwidth}
	\includegraphics[width=\textwidth]{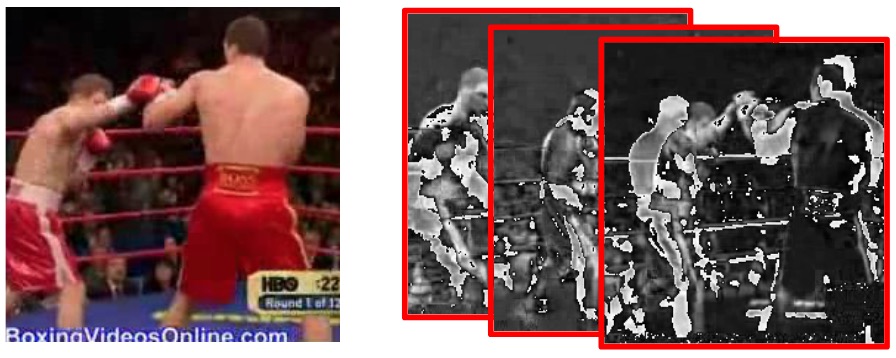}
	\caption{Class II}
	\label{class_2}
\end{subfigure}
	\begin{subfigure}[b]{0.2\textwidth}
		\includegraphics[width=\textwidth]{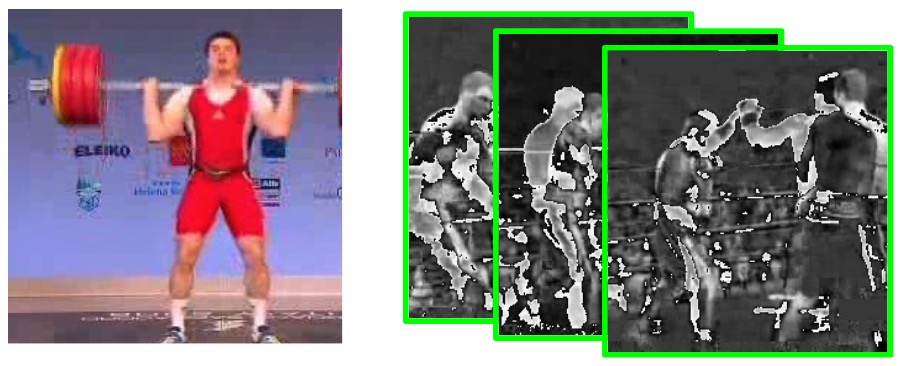}
		\caption{Class III}
		\label{class_3}
	\end{subfigure} ~
	\begin{subfigure}[b]{0.2\textwidth}
		\includegraphics[width=\textwidth]{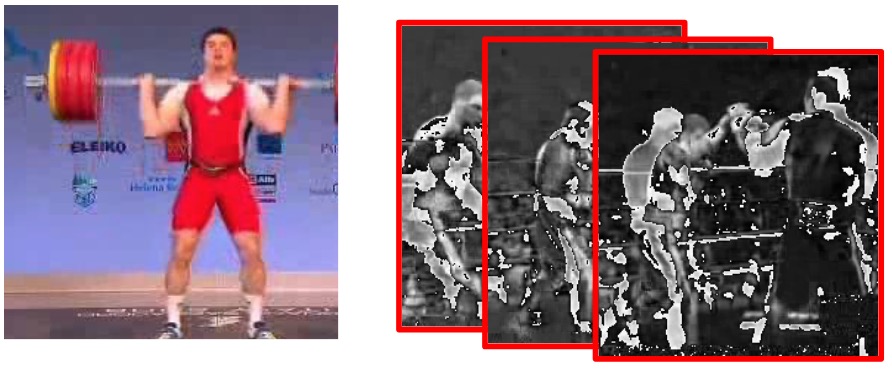}
		\caption{Class IV}
		\label{class_4}
\end{subfigure}
\caption{Self-supervised learning framework formulated as a four-class classification problem. Given a tuple of a RGB frame and a stack of difference (SOD), the network reasons about frame ordering and spatio-temporal correspondence. Valid and invalid ordered motion are highlighted in green and red respectively. Class III shows a tuple of a weight lifting RGB frame and a SOD encoding a boxing action with a valid sequence -- no spatio-temporal correspondence.}
	\label{fig:unsupervised_training}
\end{figure}

The two-stream architecture is prominent for supervised action recognition. The spatial tower performs action recognition from still video frames, whilst the temporal tower is trained to recognize action from motion in the form of dense optical flow. We propose a training scheme to leverage the two-stream architecture for self-supervised learning. Figure~\ref{fig:networkarch} shows our two-stream architecture adopting the stack of differences motion encoding and subtraction fusion from~\cite{fernando2017self}.

Recent work~\cite{misra2016shuffle,fernando2017self,lee2017unsupervised} focuses on a sequential verification task where the network reasons about the frames ordering. Our proposed framework builds on that to enforce combined spatio-temporal learning. The self-supervised learning problem is formulated as a four-classes classification task. Given a RGB frame and a motion encoding, the network answers two binary questions. Is the encoded motion sequentially valid? Does the encoded motion correspond to the spatial image representation? Figure~\ref{fig:unsupervised_training} shows four possible  tuples -- four classes.

To train the network, clips spanning 70 frames ($\approx 2.5$ seconds) are randomly sampled from videos. The center RGB frame is fed into the spatial tower. Motion, across six sequential equi-distant frames, encoded using stack of difference is fed into the motion tower. Both towers employ the AlexNet architecture. Yet, the motion tower input is a five channels image. In all experiments, the spatial tower weights are fixed to the ImageNet pre-trained weights.

To evaluated the self-supervised approach effectiveness, the self-supervised pre-trained network is fine-tuned on three datasets for action recognition and compared against random weights initialization. The HMDB51 and UCF101 are benchmark action recognition datasets that pose a limited annotations challenge. Honda driving dataset (HDD)~\cite{RamanishkaCVPR2018} is an event recognition dataset of realistic ego-motion videos for  driver behavior understanding and causal reasoning. While bigger in terms of annotations, the HDD event class distribution is long-tail which poses an imbalanced data challenge. 

	Table~\ref{tbl:eval} shows the performance of both self-supervised and supervised approaches. The self-supervised approach is consistently superior on all datasets. UCF101 is less challenging compared to HMDB51. This explains the bigger performance gain in classification accuracy.
Unlike the supervised (Rand Init.) approach, the self-supervised one is less biased against HDD small frequency events -- merge or right lane branch. This emphasizes the self-supervised approach potential to address unbalanced data problem.

\begin{table}
	\small
	\centering
\begin{tabular}{|c|c|c|}
	\hline 
	& Sup (Rand Init. )& Self-Sup \\ 
	\hline 
	HMDB51 - Split 1 & 31.47 & \textbf{33.54} \\ 
	\hline 
	UCF101 - Split 1& 52.14 & \textbf{56.25} \\ 
	\hline 
	Honda (HDD) & 81.72 & \textbf{82.78} \\ 
	\hline 
\end{tabular}  
\caption{Self-supervised mean accuracy evaluation against supervised approach initialized with random weights.}
\label{tbl:eval}
\end{table}

\section{Discussion}
\begin{figure}
	\centering
	\includegraphics[width=0.5\linewidth]{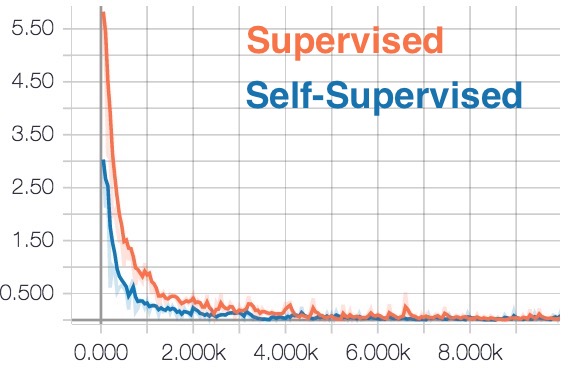}
	\caption{Training loss curves after 10K Iterations for both supervised and self-supervised approaches.}
	\label{fig:losscurve}
\end{figure}
In table~\ref{tbl:eval}, the self-supervised approach performance gains are marginal. Yet, the  preliminary results consistency across multiple datasets are promising and demand further experiments. For example, the powerful ImageNet pre-trained spatial tower quickly over fits on small datasets as shown in figure~\ref{fig:losscurve}. Thus, the network weights learning halts at early stage. Experiments to reduce overfitting and its effect on the performance gap ought to be studied

Previous self-supervised approaches count on motion representation for action recognition. Our approach uses both spatial and motion representations. Thus, straight-forward evaluation against them is biased in our favor. Experiments to separate the learned motion weights are required to fairly compare with previous work.

{\small
\bibliographystyle{ieee}
\bibliography{egbib}
}

\end{document}